\documentclass[runningheads]{llncs}

\usepackage{eccv}

\usepackage{eccvabbrv}

\usepackage{graphicx}
\usepackage{booktabs}
\usepackage{multirow}
\usepackage{colortbl}

\usepackage{hyperref}

\begin{document}

% ---------------------------------------------------------------
% TODO REVIEW: Replace with your title
\title{Dynamic Guidance Adversarial Distillation with Enhanced Teacher Knowledge} 

% TODO REVIEW: If the paper title is too long for the running head, you can set
% an abbreviated paper title here. If not, comment out.
\titlerunning{Dynamic Guidance Adversarial Distillation (DGAD)}

% TODO FINAL: Replace with your author list. 
% Include the authors' OCRID for the camera-ready version, if at all possible.
% \author{Hyejin Park\inst{1}\orcidlink{0009-0000-7258-0153} \and
% Dongbo Min\inst{1}\thanks{Corresponding author.}\orcidlink{0000-0003-4825-5240}}
\author{Hyejin Park \and
Dongbo Min\thanks{Corresponding author.}}

% TODO FINAL: Replace with an abbreviated list of authors.
\authorrunning{H. Park et al.}
% First names are abbreviated in the running head.
% If there are more than two authors, 'et al.' is used.

% TODO FINAL: Replace with your institution list.
\institute{Ewha Womans University, Seoul, South Korea \\
\email{clrara@ewha.ac.kr, dbmin@ewha.ac.kr}}

\maketitle

\begin{abstract}
  In the realm of Adversarial Distillation (AD), strategic and precise knowledge transfer from an adversarially robust teacher model to a less robust student model is paramount. Our Dynamic Guidance Adversarial Distillation (DGAD) framework directly tackles the challenge of differential sample importance, with a keen focus on rectifying the teacher model's misclassifications. DGAD employs Misclassification-Aware Partitioning (MAP) to dynamically tailor the distillation focus, optimizing the learning process by steering towards the most reliable teacher predictions. Additionally, our Error-corrective Label Swapping (ELS) corrects misclassifications of the teacher on both clean and adversarially perturbed inputs, refining the quality of knowledge transfer. Further, Predictive Consistency Regularization (PCR) guarantees consistent performance of the student model across both clean and adversarial inputs, significantly enhancing its overall robustness. By integrating these methodologies, DGAD significantly improves upon the accuracy of clean data and fortifies the model's defenses against sophisticated adversarial threats. Our experimental validation on CIFAR10, CIFAR100, and Tiny ImageNet datasets, employing various model architectures, demonstrates the efficacy of DGAD, establishing it as a promising approach for enhancing both the robustness and accuracy of student models in adversarial settings. The code is available at \href{https://github.com/kunsaram01/DGAD}{https://github.com/kunsaram01/DGAD}.
\end{abstract}

\section{Introduction}
\label{sec:intro}
Deep Neural Networks (DNNs) have significantly advanced the frontiers of image classification \cite{krizhevsky2012imagenet, he2016deep}, speech recognition \cite{graves2013speech, wang2017residual}, and natural language processing \cite{vaswani2017attention, devlin2018bert}, demonstrating remarkable success across a spectrum of complex tasks. Despite these advancements, their susceptibility to adversarial attacks \cite{szegedy2013intriguing, goodfellow2014explaining} poses a critical challenge, particularly in safety-sensitive domains such as autonomous vehicles \cite{eykholt2018robust, sobh2021adversarial} and medical diagnostics \cite{ma2021understanding, hirano2021universal}. This vulnerability becomes even more pronounced in lightweight models designed for resource-constrained environments, where their limited capacity undermines robustness. 

\begin{figure}[t!]
\begin{center}
\includegraphics[width=0.9\linewidth]{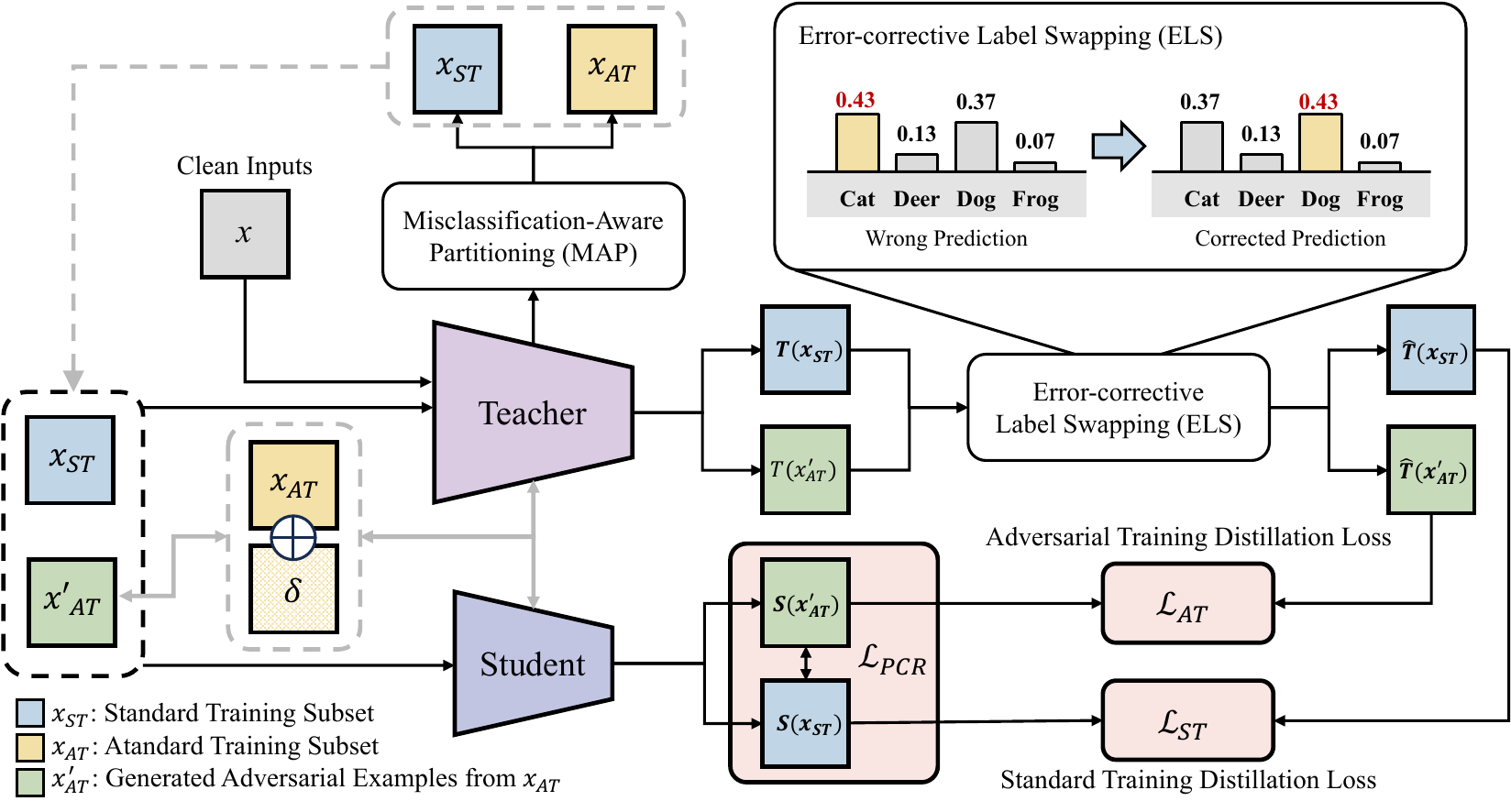}
\end{center}
\caption{\textbf{The overview of Dynamic Guidance Adversarial Distillation (DGAD)
framework.} The DGAD framework refines adversarial distillation by employing a strategic approach: Misclassification-Aware Partitioning (MAP) categorizes inputs for tailored learning, Error-corrective Label Swapping (ELS) fixes teacher’s mispredictions, and Predictive Consistency Regularization (PCR) maintains learning uniformity. Together, these methods improve student model accuracy and robustness. $S(\cdot)$ and $T(\cdot)$ are the predictions of the student and teacher models, while $\hat{T}(\cdot)$ is the corrected teacher predictions after ELS.}  
\label{fig:figure1} 
\end{figure}

Adversarial Training (AT) \cite{madry2017towards, zhang2019theoretically, pang2022robustness} has emerged as a crucial strategy to enhance the resilience of DNNs against adversarial attacks by training with adversarial examples. Although effective, the benefits of AT are more pronounced in larger models, leaving smaller models vulnerable due to their reduced capacity to handle adversarial perturbations. This limitation has led to the exploration of Adversarial Distillation (AD) \cite{goldblum2020adversarially, zhu2021reliable, huang2023boosting, maroto2022benefits} as a method to transfer the robustness and accuracy from a larger, well-trained robust teacher model to a smaller, less robust student model, aiming to bridge the performance gap under adversarial conditions.

An often-overlooked issue in AD is the direct transfer of knowledge from the teacher to the student model without addressing potential inaccuracies in the predictions of the teacher. This oversight can significantly compromise the robustness and accuracy of the student model. In response to this challenge, recent advancements in AT have developed distinct treatment of samples according to their classification status. Methods such as Margin Maximization \cite{ding2018mma, cheng2020cat} and Misclassification-Aware \cite{wang2019improving, altinisik2023a3t} strategies have demonstrated that an indiscriminate approach—particularly using adversarial examples generated from misclassified clean inputs—can decrease model robustness. These findings underscore the necessity for AD to adopt a more thoughtful and strategic approach to knowledge transfer, specifically focusing on correcting teacher errors to effectively enhance the adversarial resilience of the student model.

In this study, we introduce the \textbf{Dynamic Guidance Adversarial Distillation (DGAD)} framework (see \cref{fig:figure1}), embodying the principle of `dynamic guidance'. This concept transcends the traditional static approach to weighting distillation processes for clean and adversarial inputs by employing dynamic weighting to optimize the distillation focus. Dynamic guidance entails the real-time recognition and partitioning of training inputs within a batch, based on the teacher model's misclassification status of clean inputs. It is followed by the immediate correction of any misclassified labels for both segregated clean and adversarial inputs during distillation. By pinpointing and separating misclassified samples, DGAD enables a custom distillation strategy that optimally addresses both standard and adversarial training needs. We employ three key interventions within this framework to ensure the precise and effective transfer of knowledge to the student model: 
1) \textbf{Misclassification-Aware Partitioning (MAP)}: To realize dynamic weighting, this strategy separates the training dataset into two subsets based on the prediction of the teacher on clean inputs—The \textit{\textbf{S}tandard \textbf{T}raining (ST)} subset comprises clean inputs incorrectly classified by the teacher, emphasizing correction of these misclassifications during standard training. Conversely, the \textit{\textbf{A}dversarial \textbf{T}raining (AT)} subset includes clean inputs correctly classified by the teacher, using adversarially perturbed versions of these inputs to increase the resistance of the student model to adversarial attacks.
2) \textbf{Error-corrective Label Swapping (ELS)}: Building upon the MAP, ELS is applied to inputs where the teacher's predictions remain incorrect, specifically including the ST subset and adversarial examples generated using the AT subset. By replacing the incorrect labels predicted by the teacher with the correct ones, ELS ensures that the student model learns from accurate labels, directly addressing and amending the teacher's prediction errors observed during distillation. 
3) \textbf{Predictive Consistency Regularization (PCR)}: PCR addresses the imbalance between standard and adversarial training caused by the separate learning of ST and AT subsets in MAP. By regularizing the prediction consistency of the student model across the entire dataset, PCR ensures consistent predictions for both original inputs and their adversarial examples. This approach maintains balanced and effective learning, preventing biases toward any specific subset. 
%PCR addresses the imbalance between ST and AT, arising from learning exclusively from distinct subsets created during the MAP process. By promoting consistent predictions across the entirety of the data, PCR ensures the student model does not develop biases toward either subset, maintaining balanced and effective learning across all inputs.

By integrating these innovative strategies, DGAD transcends traditional distillation enhancements, dynamically rectifying teacher model inaccuracies while fine-tuning the knowledge transfer. This dual-action approach not only elevates the student model's defense against adversarial attacks but also significantly boosts its precision on clean data, setting a new standard for both robustness and accuracy in adversarial distillation.

\section{Related Work}
\subsection{Adversarial Training}
Adversarial Training (AT) \cite{goodfellow2014explaining, madry2017towards} is a defensive strategy against adversarial attacks, which aim to deceive machine learning models with subtly altered inputs. The central goal of AT is to train models to accurately classify these manipulated inputs. However, treating all adversarially perturbed examples with the same target labels can lead to overfitting these adversarial examples. To address the trade-off between accuracy and robustness, approaches like Adversarial Logit Pairing (ALP)\cite{kannan2018adversarial} focus on maintaining consistency between the logits of original and adversarial examples, while TRADES\cite{zhang2019theoretically} and SCORE\cite{pang2022robustness} introduce surrogate loss based on the Kullback-Leibler divergence and Squared Error loss, respectively, between the probability distributions of original and adversarial inputs.

Despite these advancements, previous research often overlooked whether adversarial examples were generated from correctly classified clean inputs. It has been highlighted that generating adversarial examples from misclassified images can exacerbate overfitting to adversarial examples. In response, methods such as MMA \cite{ding2018mma} and Misclassification-Aware Adversarial Training (MART) \cite{wang2019improving} suggest adjusting the weight of the adversarial perturbation or the loss function during adversarial training based on the misclassification of samples. These proposals underscore the importance of distinguishing between correctly classified and misclassified samples in generating adversarial examples, aiming to improve model robustness without compromising the model's ability to generalize. 

\subsection{Adversarial Distillation}
Adversarial Distillation (AD) emerged from the desire to convey the adversarial robustness of a well-trained teacher model to more compact student model. Adversarial Robust Distillation (ARD) \cite{goldblum2020adversarially} pioneered this realm by integrating Knowledge Distillation \cite{hinton2015distilling} with Adversarial Training. RSLAD \cite{zi2021revisiting} emphasized the significance of using robust soft labels in the inner optimization to generate adversarial examples. AdaAD \cite{huang2023boosting} further refined this approach, optimizing the adversarial example generation to account for discrepancies between teacher and student predictions and leveraging these refined adversarial examples for more effective training. Introspective Adversarial Distillation (IAD) \cite{zhu2021reliable} addresses teacher's unreliability in later training stages by incorporating a partial reliance on teacher's predictions, increasingly favoring the student's self-derived knowledge as training advances.

\section{Preliminaries}
\label{sec:prelim}
There is active exploration into machine learning models based on the adversarial distillation (AD) that appropriately balance accuracy and resilience to adversarial attacks. Central to this challenge is an effective transfer of knowledge from an adversarially trained teacher model to a student model, aiming to instill both accuracy and robustness.

The foundation of our investigation is knowledge distillation (KD) \cite{hinton2015distilling}, where a smaller student model is trained to mimic a more complex teacher model by aligning its predictions with those of the teacher, following the objective function in \cref{eq:kd}:
\begin{equation}
\underset{\theta}{\text{argmin}} \, (1-\alpha) \cdot \mathcal{CE}(S_{\theta}(x), y) + \alpha \cdot \tau^2 \cdot \mathcal{KL}(S_{\theta}^{\tau}(x) \, || \, T^{\tau}(x))
\label{eq:kd}
\end{equation}
where $\mathcal{CE}$ is the cross-entropy loss assessing the accuracy for an input $x$ of the student with a ground truth $y$, $\mathcal{KL}$ measures the disparity between the softened outputs of the student $S_{\theta}^{\tau}(x)$ with learnable parameters $\theta$ and a pretrained teacher model $T^{\tau}(x)$ modulated by a temperature parameter $\tau$, and $\alpha$ weights the importance of classification accuracy versus prediction similarity. 

Adversarial Robustness Distillation (ARD) \cite{goldblum2020adversarially} formulates the adversarial training in KD framework, harnessing the insights from a pretrained teacher model to guide a student model through adversarial scenarios. In contrast to AT, they use the predictions from the teacher model as reference signals. These signals aid the learning process of the student, encompassing both clean and adversarially perturbed inputs. AD adopts a min-max optimization framework that is similar to AT but is distinctively enhanced by the knowledge of the teacher model. The AD process is captured by the following optimization function: 
\begin{equation}
\begin{aligned}
&\underset{\theta}{\text{argmin}} (1-\alpha) \cdot \mathcal{CE}(S(x), y) + \alpha \cdot \mathcal{KL}(S(x'), T(x)) \\
&\text{where} \quad x' = \underset{||\delta||_{p} < \epsilon}{\mathrm{argmax}} \, \mathcal{CE}(S_{\theta}(x+\delta),y).
\end{aligned}
\label{eq:ard}
\end{equation}

Robust Soft Label Adversarial Distillation (RSLAD) \cite{zi2021revisiting} showcases the use of robust soft labels, generated by a larger, robust teacher model, to guide the student model's training on both clean and adversarial examples. This approach includes generating adversarial examples that leverage these robust soft labels for an enhanced training process. They apply robust soft labels in both of two processes: 
\begin{equation}
\begin{aligned}
&\underset{\theta}{\text{argmin}} \, (1-\alpha) \cdot \mathcal{KL}(S_{\theta}(x)||T(x)) + \alpha \cdot \mathcal{KL}(S_{\theta}(x')||T(x)), \\
&\text{where} \quad x' = \underset{||\delta||_{p} < \epsilon}{\mathrm{argmax}} \, \mathcal{KL}(S_{\theta}(x+\delta),T(x)).
\end{aligned}
\label{eq:rslad}
\end{equation}

In \cref{eq:rslad}, the prediction deviation between the student and teacher models is assessed using both a clean input $x$ and its corresponding adversarial example $x'$. The adversarial example is produced during an inner-maximization phase, where a deliberate perturbation $\delta$ is applied to the clean input within an $\epsilon$-constrained sphere to maximize the divergence and hence challenge the model. The outer-minimization phase then involves the student model training, thereby bolstering the resilience of model to adversarial perturbations and mirroring robust predictive qualities of the teacher model. 

Adaptive Adversarial Distillation (AdaAD) \cite{huang2023boosting} further enhances adversarial distillation by integrating the teacher model in the generation of adversarial examples and guiding the student model with well-estimated probabilities for each data point and its $\epsilon$-neighborhood region. This approach mitigates model over-smoothness, thereby reducing the adversarial trade-offs for enhanced generalization. The AdaAD objective, shown in \cref{eq:adaad}, emphasizes the teacher-directed adversarial learning:
\begin{equation}
\begin{aligned}
&\underset{\theta}{\text{argmin}} \, (1-\alpha) \cdot \mathcal{KL}(S_{\theta}(x)||T(x)) + \alpha \cdot \mathcal{KL}(S_{\theta}(x')||T(x')), \\
&\text{where} \quad x' = \underset{||\delta||_{p} < \epsilon}{\mathrm{argmax}} \, \mathcal{KL}(S_{\theta}(x+\delta),T(x+\delta)).
\end{aligned}
\label{eq:adaad}
\end{equation}

A key advance lies in the inner-maximization process, which creates adversarial examples that maximize the discrepancy between the predictions of the student and teacher models on adversarially perturbed inputs. The teacher predictions on these adversarial examples are then used as supervisory signals to guide the training of the student model.

% To sum up, the evolution from KD to AD and now to AdaAD, highlights significant strides in developing models that maintain accuracy in the face of adversarial threats while avoiding over-specialization on training data. AdaAD, with its nuanced approach to adversarial example generation, emerges as a cutting-edge method in adversarial distillation for compact models.

\section{Dynamic Guidance Adversarial Distillation}
\label{sec:method}

\subsection{Motivation of DGAD}
\begin{figure*}[t!]
\begin{center}
\includegraphics[width=0.99\linewidth]{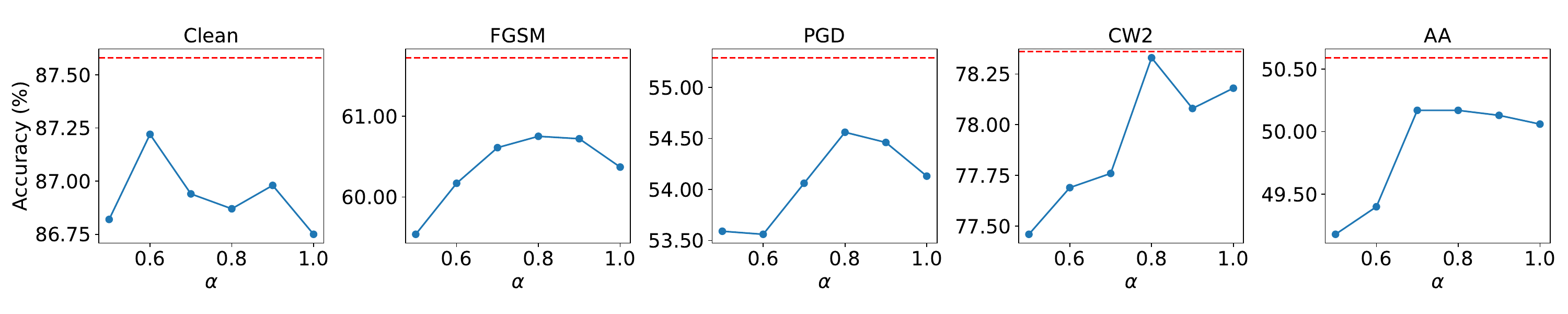}
\end{center}
\caption{\textbf{Necessity of dynamically varying AD loss weights for individual samples. } We compare the performance of AdaAD \cite{huang2023boosting}, which originally proposes to employ a static weight $\alpha$ in (\cref{eq:adaad}), against our Dynamic Guidance Adversarial Distillation (DGAD) that adapts the weight dynamically per sample as in \cref{eq:map}. To validate the importance of dynamical weights, we adjust $\alpha$ for AdaAD and compare it across clean and adversarial scenarios. The blue solid line represents AdaAD's performance with a fixed $\alpha$ across all samples, while the red dotted line indicates DGAD's performance, showing improved accuracy due to its dynamic weighting approach.}  
\label{fig:figure2} 
\end{figure*}

In the realm of Adversarial Distillation (AD), reliance on static weighting for loss across all samples, notably in frameworks like AdaAD \cite{huang2023boosting} and similar approaches \cite{goldblum2020adversarially, maroto2022benefits, zhu2021reliable, zi2021revisiting}, often results in an imbalance between maintaining accuracy on original inputs and ensuring robustness against adversarial threats. This issue becomes more pronounced when adjusting the weighting parameter $\alpha$, as depicted in \cref{fig:figure2}. Static weights fail to account for the varying importance of individual samples, leading to a suboptimal balance between accuracy and robustness. This lack of consideration for sample importance means that some samples, particularly those misclassified by the teacher model, are not properly weighted during training. Consequently, the inaccuracies of the teacher model can disproportionately influence the training process, propagating these errors to the student model and undermining the overall effectiveness of the distillation process.

Our Dynamic Guidance Adversarial Distillation (DGAD) framework addresses this problem through dynamic adjustment of weighting, tailored to the precision of the teacher model's predictions. DGAD hinges on a critical insight: the importance of each sample should be dynamically adjusted based on the accuracy of the teacher model's predictions on clean inputs. When the teacher model's predictions on clean inputs are incorrect, using these misclassified samples to generate adversarial examples can degrade the student's learning experience during distillation. To mitigate this, DGAD dynamically adjusts the weights assigned to each sample based on whether the teacher model's prediction for clean inputs is correct. By deliberately excluding misclassified samples from the adversarial generation process and focusing adversarial training on accurately classified samples, DGAD ensures that the student model's training benefits from the most reliable information. For samples misclassified by the teacher model, the focus is on improving the student's performance on clean data. This methodological pivot enhances the student model's resilience to adversarial manipulations while either maintaining or improving accuracy on clean data, thereby strengthening the model's overall performance.

The motivation behind DGAD is to navigate and mitigate the intrinsic trade-offs present in AD, employing a dynamic and discerning strategy for knowledge transfer that focuses exclusively on transmitting dependable insights from the teacher model. The subsequent sections will delve deeper into the strategies that embody this approach—Misclassification-Aware Partitioning (MAP), Error-corrective Label Swapping (ELS), and Predictive Consistency Regularization (PCR)—showcasing DGAD's commitment to achieving both effectiveness and efficiency in enhancing model robustness and accuracy.

\subsection{Misclassification-Aware Partitioning in AD} 

Dynamic sample weighting is essential for effective knowledge distillation. Our DGAD framework implements this through the Misclassification-Aware Partitioning (MAP) strategy, which categorizes the dataset into two subsets for specialized training:
(1) \textbf{Standard Training Subset ($x_{ST}$)} consists of samples where the teacher model’s predictions for clean inputs $x$ are incorrect, denoted as $x_{\scriptscriptstyle ST} = \{x \mid \arg\max(T(x)) \neq y\}$. These samples are used in the standard distillation path to improve the student's accuracy on clean data; (2) \textbf{Adversarial Training Subset ($x_{AT}$)} consists of samples correctly classified by the teacher, represented as $x_{\scriptscriptstyle AT} = \{x \mid \arg\max(T(x)) = y\}$. These samples are used for adversarial distillation to enhance the student model's robustness against adversarial perturbations.

This partitioning allows for targeted distillation, optimizing the contribution of each sample to the student's learning process. The training objective combines standard and adversarial distillation to balance and streamline the learning process:
\begin{equation}
\begin{aligned}
\underset{\theta}{\text{argmin}} \, &\underbrace{\mathcal{KL}(S_{\theta}(x_{\scriptscriptstyle ST})||T(x_{\scriptscriptstyle ST}))}_{\text{Standard Training Distillation $\mathcal{L}_{ST}$}} + \underbrace{\mathcal{KL}(S_{\theta}(x_{\scriptscriptstyle AT}')||T(x_{\scriptscriptstyle AT}'))}_{\text{Adversarial Training Distillation $\mathcal{L}_{AT}$}} \\
& \text{where} \quad x_{\scriptscriptstyle AT}' = \underset{||\delta||_{p} < \epsilon}{\mathrm{argmax}} \, \mathcal{KL}(S_{\theta}(x_{\scriptscriptstyle AT}+\delta)||T(x_{\scriptscriptstyle AT}+\delta)).
\end{aligned}
\label{eq:map}
\end{equation}
where $x_{\scriptscriptstyle AT}'$ represents adversarially perturbed inputs from $x_{\scriptscriptstyle AT}$. 

\textbf{Standard Training Distillation ($\mathcal{L}_{ST}$):} Aims to minimize the divergence between the student and teacher model predictions for the standard training subset $x_{\scriptscriptstyle ST}$, comprising samples inaccurately classified by the teacher. This focuses the distillation on enhancing the student's accuracy on clean data, ensuring the student model learns more precisely from foundational data, contributing to overall performance improvements.

\textbf{Adversarial Training Distillation ($\mathcal{L}_{AT}$):} Targets adversarial resilience by distilling knowledge from adversarially perturbed examples $x_{\scriptscriptstyle AT}'$, derived from samples correctly identified by the teacher. This approach supports the student model in maintaining robustness in adversarial situations.

MAP’s approach of generating adversarial examples from accurately classified samples avoids propagating teacher model inaccuracies. This precision in knowledge transfer, as our ablation study (\cref{tab:ablation1}) demonstrates, substantially boosts the learning dynamics of the student model, marking a significant advance in adversarial distillation efficacy.

\subsection{Error-corrective Label Swapping}

Error-corrective Label Swapping (ELS) is a pivotal strategy designed to rectify inaccuracies in the predictions of the teacher model, especially focusing on samples misclassified after implementing MAP. ELS comes into play when a discrepancy is identified—specifically, when the teacher model wrongly places higher confidence in an incorrect label $\hat{y}$ over the correct label $y$. This discrepancy is measured through a negative margin $M$, which triggers the corrective mechanism of ELS. By swapping the labels in such instances, ELS ensures that the student model receives and learns from correct labels, enhancing the precision and reliability of knowledge transfer. This corrective action is crucial for two scenarios.

All samples in \textbf{clean inputs $x_{\scriptscriptstyle ST}$} undergo label swapping to rectify the teacher's initial misclassifications, ensuring $x_{\scriptscriptstyle ST}$ contributes positively to the student's learning. Here, $y$ is a true label and $\hat{y}$ is a predicted label, $P_{T}$ is a softmax probability of the teacher model, and $\hat{P}_{T}$ is the adjusted probability after swapping the incorrect prediction with the correct label:
\begin{equation}
\begin{aligned}
\hat{P}_{T} \leftarrow \mathcal{SWAP} (P_{T}(\hat{y}|x_{\scriptscriptstyle ST}), P_{T}(y|x_{\scriptscriptstyle ST})), \quad \forall x_{\scriptscriptstyle ST}.
\end{aligned}
\label{eq:ls_st}
\end{equation} 

\textbf{Adversarial examples $x_{\scriptscriptstyle AT}'$} are generated based on the student model using $x_{\scriptscriptstyle AT}$. According to IAD \cite{zhu2021reliable}, the teacher's predictions on $x_{\scriptscriptstyle AT}'$ may become unreliable as student model training progresses. To prevent the propagation of these unreliable predictions during later stages of training, ELS is applied only when the teacher's predictions on $x'_{\scriptscriptstyle AT}$ are incorrect. This ensures that the adversarial training of the student model is based on accurate teacher feedback: 
% To prevent the propagation of these errors to the student model, conditional label swapping corrects the mispredictions of the teacher model on $x'_{\scriptscriptstyle AT}$, ensuring that the adversarial training of the student model is based on corrected teacher feedback:
\begin{equation}
\begin{aligned}
\hat{P}_{T} &\leftarrow \mathcal{SWAP} (P_{T}(\hat{y}|x'_{\scriptscriptstyle AT}), P_{T}(y|x'_{\scriptscriptstyle AT})), \quad \text{if } M < 0,\\
\text{where} &\quad M = P_{T}(y|x'_{\scriptscriptstyle AT}) - P_{T}(\hat{y}|x'_{\scriptscriptstyle AT}), \quad \text{for generated } x'_{\scriptscriptstyle AT}.
\end{aligned}
\label{eq:ls_at}
\end{equation} 

By systematically correcting these errors, ELS substantially enhances the quality of knowledge distilled to the student model and ensures a more effective and accurate learning process. This strategy is instrumental in overcoming the limitations posed by misclassifications, significantly contributing to the robustness and accuracy of the student model as demonstrated in our subsequent ablation studies. The training objectives, $\mathcal{L}_{ST}$ for standard inputs and $\mathcal{L}_{AT}$ for adversarial inputs, are refined through corrected teacher predictions $\widehat{T}(\cdot)$ to ensure an optimal distillation path.
\begin{equation}
\begin{aligned}
\underset{\theta}{\text{argmin}} \, \underbrace{\mathcal{KL}(S_{\theta}(x_{\scriptscriptstyle ST})||\widehat{T}(x_{\scriptscriptstyle ST}))}_{\text{Standard Training Distillation $\mathcal{L}_{ST}$}} + \underbrace{\mathcal{KL}(S_{\theta}(x_{\scriptscriptstyle AT}')||\widehat{T}(x_{\scriptscriptstyle AT}')).}_{\text{Adversarial Training Distillation $\mathcal{L}_{AT}$}}
\end{aligned}
\label{eq:els}
\end{equation}

\subsection{Predictive Consistency Regularization} 

Predictive Consistency Regularization (PCR) directly addresses the challenge of maintaining consistency in the student model's predictions across both Standard Training (ST) and Adversarial Training (AT) subsets. Given that ST focuses on correcting misclassifications of clean inputs and AT concentrates on enhancing resilience against adversarial perturbations, an inherent risk emerges: the student model might develop inconsistent responses to similar inputs under different contexts. PCR works to bridge this gap, ensuring that the student model applies a consistent learning approach to both subsets. By doing so, PCR mitigates the potential for divergent behaviors, fostering a unified model performance regardless of the input's nature—clean or adversarially perturbed.

PCR introduced via $\mathcal{L}_{\scriptscriptstyle PCR}$, harmonizes the student model's responses to clean ($x$) and their corresponding adversarial ($x'$) inputs. This regularization approach is instrumental in fostering a balanced learning process, as evidenced by our ablation study results in \cref{tab:ablation1}. Here, $\mathcal{L}_{\scriptscriptstyle PCR} = ||S_{\theta}(x) - S_{\theta}(x')||_{2}$. The comprehensive approach to adversarial distillation is encapsulated in the total loss $\mathcal{L}_{DGAD}$, defined as follows:
\begin{equation}
\begin{aligned}
\mathcal{L}_{DGAD} = \mathcal{L}_{ST} + \mathcal{L}_{AT}+ \beta \cdot \mathcal{L}_{PCR},
\end{aligned}
\label{eq:total}
\end{equation}
This loss function strategically emphasizes predictive consistency through the parameter $\beta$, enhancing the student model's accuracy and robustness in a comprehensive manner.

Implementing insights from AT research \cite{zhang2019theoretically, pang2022robustness}, PCR distinctly tailors these principles to our framework in AD, achieving a strategic balance between accuracy and adversarial resilience. Along with MAP and ELS, significantly elevates defense mechanisms against adversarial threats, a claim substantiated by our ablation study's robust performance enhancements against a range of attacks.

\begin{table}[t!]
\caption{\textbf{Performance of Teacher Models on CIFAR10/ CIFAR100 and TinyImageNet.}}
\centering
\scriptsize\addtolength{\tabcolsep}{6.5pt}
\begin{tabular}{l|l|c|c|c|c|c}
\toprule
Dataset & Teacher & Clean & FGSM & PGD & CW & AA \\
\midrule
CIFAR10 & ResNet18 \cite{zhang2019theoretically} & 82.94 & 59.02 & 53.71 & 77.04 & 49.34 \\
CIFAR10 & WideResNet-34-10 \cite{pang2020bag} & 87.20 & 62.14 & 55.90 & 77.80 & 51.79 \\
CIFAR10 & WideResNet-34-20 \cite{chen2021ltd} & 86.03 & 66.01 & 63.33 & 82.60 & 57.71 \\
CIFAR100 & WideResNet-34-10 \cite{chen2021ltd} & 64.07 & 39.83 & 36.61 & 56.22 & 30.57 \\
Tiny ImageNet & PreActResNet18 \cite{he2016identity} & 46.04 & 22.36 & 20.85 &	41.00 &	15.45 \\
\bottomrule
\end{tabular}
\label{tab:teacher_performance}
\end{table}

\section{Experiments}
\textbf{Experimental Setup.} The performance of DGAD was assessed on the CIFAR10, CIFAR100 \cite{krizhevsky2009learning}, Tiny ImageNet \cite{le2015tiny} datasets, normalized between [0,1]. Benchmarks included PGD-AT \cite{madry2017towards}, TRADES \cite{zhang2019theoretically}, and several AD methods (ARD \cite{goldblum2020adversarially}, IAD \cite{zhu2021reliable}, RSLAD \cite{zi2021revisiting}, AKD \cite{maroto2022benefits}, AdaAD \cite{huang2023boosting}). We employed ResNet18 \cite{he2016deep} and MobileNetV2 \cite{sandler2018mobilenetv2} as students, and WideResNet-34-10 (both datasets), WideResNet-34-20 (CIFAR10) \cite{pang2020bag, chen2021ltd}, PreActResNet18 \cite{he2016identity} (Tiny ImageNet) as teachers. \cref{tab:teacher_performance} provides the performance of the teacher models used in our experiments.
For fair comparison, models were trained following AdaAD's basic settings. We used SGD with an initial learning rate of 0.1, momentum of 0.9, weight decay of 5e-4, and standard data augmentation. Training duration varied: PGD-AT stopped at 110 epochs, TRADES and AD methods \cite{goldblum2020adversarially, zhu2021reliable, zi2021revisiting, maroto2022benefits}, including DGAD, ran for 200 epochs with learning rate adjustments at epochs 100 and 150. Inner optimization parameters for adversarial training included 10 iterations, a step size of 2/255, and a perturbation bound of 8/255 under $L_{\infty}$ constraint. Hyper-parameters $\alpha$ and distillation temperature $\tau$ were set as recommended. For the loss function, $\mathcal{L}_{\scriptscriptstyle PCR}$ weight $\beta$ was set to 5 for ResNet18, 10 for MobileNetV2, and 15 for PreActResNet18 models. Experiments were conducted in PyTorch with an adversarial training library.

\noindent\textbf{Evaluation Metrics.} 
Model performance is gauged through natural accuracy on clean samples and robust accuracy against adversarial samples, tested using FGSM \cite{goodfellow2014explaining}, PGD \cite{madry2017towards}, CW2 \cite{carlini2017towards}, and AutoAttack (AA) \cite{croce2020reliable}. Perturbation size for FGSM, PGD, and AA is set at 8/255, with PGD utilizing 10 steps of 2/255 each. CW2's equilibrium constant is 0.1. Results reflect the best PGD-10 checkpoint.

\begin{table}[t!]
\caption{\textbf{Efficacy of DGAD Components on CIFAR10.} We utilize ResNet18 (student) and WideResNet-34-10 (teacher) to test components including Misclassification-Aware Partitioning (MAP), Error-corrective Label Swapping (ELS), and Predictive Consistency Regularization (PCR). Notations are as follows: $x'$ - misclassified adversarial examples without consider misclassification on clean inputs, $x_{\scriptscriptstyle ST}$ – misclassified clean inputs, $x'_{\scriptscriptstyle AT}$ – misclassified adversarial examples.}
\centering
\scriptsize\addtolength{\tabcolsep}{8.5pt}
\begin{tabular}{l|c|c|c|c|c}
\toprule
Method & Clean & FGSM & PGD & CW & AA \\
\midrule
Baseline \cite{huang2023boosting} & 86.75 & 60.37 & 54.13 & 78.18 & 50.06 \\
+MAP                       & 86.92 & 61.40 & 54.94 & 78.34 & 50.82 \\
+MAP+PCR                       & 87.19 & 61.14 & 54.92 & 78.72 & 50.52  \\
+ELS($x'$)                 & 87.27 & 60.66 & 54.47 & 78.42 & 50.13 \\
+MAP+ELS($x'_{\scriptscriptstyle AT}$)              & 87.28 & 61.48 & 54.97 & 77.85 & 50.80 \\
+MAP+ELS($x_{\scriptscriptstyle ST}$)               & 87.81 & 61.14 & 54.89 & 78.36 & 50.26 \\
+MAP+ELS($x_{\scriptscriptstyle ST}$)+ELS($x'_{\scriptscriptstyle AT}$)      & 87.53 & 61.23 & 54.77 & 78.49 & 50.33 \\
+MAP+ELS($x_{\scriptscriptstyle ST}$)+ELS($x'_{\scriptscriptstyle AT}$)+PCR  & 87.58 & 61.72 & 55.29 & 78.36 & 50.63 \\
\bottomrule
\end{tabular}
\label{tab:ablation1}
\end{table}

\begin{table}[t!]
\centering
\caption{\textbf{Impact of different labeling methods in DGAD on CIFAR10.} The experimental setup is identical to that described in \cref{tab:ablation1}.}
\scriptsize\addtolength{\tabcolsep}{14pt}
\begin{tabular}{l|c|c|c|c|c}
\toprule
Method & Clean & FGSM & PGD & CW & AA \\
\midrule
Label Smoothing & 87.24 & 61.97 & 56.09 & 77.51 & 49.86\\
Label Mixing & 87.59 & 61.90 & 55.22 & 78.64 & 50.64\\
\textbf{Label Swapping} & 87.58 & 61.72 & 55.29 & 78.36 & 50.63\\
\bottomrule
\end{tabular}
\label{tab:ablation2}
\end{table}

\subsection{Ablation Study} 
\label{sec:main_ablation}

\textbf{Efficacy of Individual Components.} 
The comprehensive ablation study presented in \cref{tab:ablation1} meticulously dissects the distinct and combined influences of the proposed components, revealing a clear trajectory of performance enhancements and robustness against adversarial threats. 

When Misclassification-Aware Partitioning (MAP) is applied independently, it yields a significant and vital enhancement in model robustness. This underscores the fundamental efficacy of MAP in directing the student model's focus towards the most reliable predictions of the teacher.

Error-corrective Label Swapping (ELS) presents its own set of advantages. When ELS is applied to adversarial examples $x'$ generated without MAP, we observe enhanced robustness and accuracy. Further improvements in robustness are noted when applying ELS on $x'_{\scriptscriptstyle AT}$ after MAP, highlighting the benefits of excluding misclassified clean inputs and the crucial role of addressing misclassifications in bolstering adversarial resilience. Using ELS in $x_{\scriptscriptstyle ST}$ with MAP amplifies performance. This demonstrates the critical role of rectifying teacher errors, as correcting misclassifications on clean data substantially boosts learning and robustness. Applying ELS to both $x_{\scriptscriptstyle ST}$ and $x'_{\scriptscriptstyle AT}$ with MAP enhances this effect, highlighting the synergy of these strategies in improving the performance of the student model.

Predictive Consistency Regularization (PCR) not only maintains robustness gains from MAP but also enhances accuracy on clean inputs, showcasing the synergistic effect of the two components. The integration of all components, including MAP and PCR, significantly outperforms configurations without PCR, highlighting the role of PCR in complementing and augmenting MAP and ELS.

The initial addition of  MAP notably improved AA performance by 0.77\%, with subsequent ELS and PCR enhancements showing smaller AA improvements. In total, these components resulted in an improvement of 0.57\% over the baseline in AA. This non-linear improvement arises because our method aims to balance clean accuracy and adversarial robustness. %The ablation study highlights the value of DGAD's components: MAP establishes robustness, ELS enhances accuracy and defense, and PCR balances robustness with accuracy. Together, they create a model that excels in both defense against adversarial threats and precision, showcasing the effectiveness of our framework.

\textbf{Effectiveness of Labeling Techniques in DGAD.} Within the DGAD, we evaluate labeling techniques for correcting teacher misclassifications. Label Swapping is compared with Label Smoothing, represented as $(1-\alpha) \cdot y + \alpha \cdot \frac{1}{C}$, where $C$ is the number of classes, and Label Mixing, shown as $\alpha \cdot T(x) + (1 - \alpha) \cdot y$ (\cref{tab:ablation2}). While Label Smoothing slightly improves clean data accuracy, its impact on robustness varies. Label Mixing and Label Swapping show similar results in enhancing accuracy and robustness. However, Label Mixing's reliance on the hyperparameter $\alpha$ can complicate corrections, especially with overly confident incorrect predictions. Label Swapping directly corrects misclassifications, simplifying the training process and ensuring precise knowledge transfer without complex parameters tuning. This highlights its advantage in adversarial training and provides a clear rationale for its use in DGAD. 

\begin{table*}[t!]
\centering
\scriptsize\addtolength{\tabcolsep}{1pt}
\caption{\textbf{Evaluating on CIFAR10.} RN-18 and MN-V2 denote the student models ResNet-18 and MobileNetV2, respectively. Best results in \textbf{bold}; next-best \underline{underlined}.} 
\label{tab:cifar10}
\begin{tabular}{c|l|c|c|c|c|c|c|c|c|c|c}
\toprule
\multicolumn{2}{c|}{Teacher Model} & \multicolumn{5}{c|}{WideResNet-34-10} & \multicolumn{5}{c}{WideResNet-34-20} \\
% \cline{1-12}
\midrule
model & method & Clean & FGSM & PGD & CW2 & AA & Clean & FGSM & PGD & CW2 & AA \\
\midrule
\multirow{9}{*}{RN-18} & PGD-AT\cite{madry2017towards} & 82.95 & 57.16 & 52.87 & 77.56 & 47.69 & 82.95 & 57.16 & 52.87 & 77.56 & 47.69\\
& TRADES\cite{zhang2019theoretically} & 83.00 & 58.42 & 53.18 & 76.92 & 49.21  & 83.00 & 58.42 & 53.18 & 76.92 & 49.21\\
& ARD\cite{goldblum2020adversarially} & 84.04 & 58.26 & 52.67 & 74.95 & 48.62  & 84.03 & 58.16 & 53.11 & 79.13 & 48.07\\
& IAD\cite{zhu2021reliable} & 83.19 & 57.76 & 53.17 & 76.77 & 48.82  & 84.71 & \underline{61.28} & 54.92 & 79.44 & 49.85\\
& RSLAD\cite{zi2021revisiting} & 83.60 & 57.45 & 52.60 & 76.85 & 48.45  & 83.52 & 58.36 & 53.46 & 78.36 & 48.66\\
& AKD\cite{maroto2022benefits} & 84.69 & 58.97 & 53.28 & 77.25 & 48.37  & 83.22 & 58.63 & 54.16 & 78.44 & 49.26\\
& AdaAD\cite{huang2023boosting} & \underline{86.75} & \underline{60.37} & \underline{54.13} & \underline{78.18} & \underline{50.06}  & \underline{85.58} & 60.85 & \underline{56.40} & \underline{80.83} & \underline{51.37}\\
& \textbf{DGAD} & \textbf{87.58} & \textbf{61.72} & \textbf{55.29} & \textbf{78.36} & \textbf{50.59} & \textbf{85.75}  & \textbf{62.28} & \textbf{58.05} & \textbf{81.60}  & \textbf{52.34} \\ 
& & +0.83 & +1.35 & +1.16 & +0.17 & +0.53  & +0.17 & +1.00 & +1.65 & +0.77 & +0.97\\
\midrule
\multirow{9}{*}{MN-V2} & PGD-AT\cite{madry2017towards}  & 77.54 & 53.58 & 49.90 & 72.54 & 44.56 & 77.54 & 53.58 & 49.90 & 72.54 & 44.56\\
& TRADES\cite{zhang2019theoretically} & 79.80 & 54.84 & 50.51 & 75.30 & 45.67  & 79.80 & 54.84 & 50.51 & 75.30 & 45.67\\
& ARD\cite{goldblum2020adversarially} & 84.63 & 58.00 & 50.82 & 72.93 & 46.48  & 79.56 & 53.17 & 49.06 & 74.51 & 44.04\\
& IAD\cite{zhu2021reliable} & 82.11 & 55.27 & 50.20 & 75.41 & 45.66  & 83.31 & 58.29 & 52.98 & 78.03 & 47.11\\
& RSLAD\cite{zi2021revisiting} & 83.24 & 56.69 & 51.57 & 76.52 & 47.18  & 81.11 & 56.39 & 51.66 & 76.20 & 46.75\\
& AKD\cite{maroto2022benefits} & 82.64 & 56.17 & 50.49 & 75.31 & 45.67  & 83.41 & \underline{57.71} & 52.35 & 77.97 & 46.82\\
& AdaAD\cite{huang2023boosting} & \underline{86.80} & \underline{58.56} & \underline{52.00} & \underline{78.27} & \underline{47.97}  & \underline{83.79} & 57.29 & \underline{53.04} & \underline{79.24} & \underline{47.66}\\
& \textbf{DGAD} & \textbf{87.19} & \textbf{60.11} & \textbf{53.56} & \textbf{79.40} & \textbf{49.19} &  \textbf{85.30} & \textbf{61.20} & \textbf{56.77} & \textbf{80.98} & \textbf{51.10} \\
&  & +0.39 & +1.55 & +1.56 & +1.13 & +1.22 & +1.51 & +3.49 & +3.73 & +1.74 & +3.44\\
\bottomrule
\end{tabular}
\end{table*}

\subsection{Adversarial Robustness}

\begin{table}[t!]
\centering
\scriptsize\addtolength{\tabcolsep}{10.2pt}
\caption{\textbf{Evaluating on CIFAR100.} RN-18 and MN-V2 denote the student models ResNet-18 and MobileNetV2, respectively. Best results in \textbf{bold}; next-best \underline{underlined}.}
\label{tab:cifar100}
\begin{tabular}{c|l|c|c|c|c|c}
\toprule
\multicolumn{2}{c|}{Teacher Model} & \multicolumn{5}{c}{WideResNet-34-10} \\
% \cline{1-7}
\midrule
Model & Method & Clean & FGSM & PGD & CW2 & AA \\
\midrule
\multirow{10}{*}{RN-18} & PGD-AT\cite{madry2017towards} & 56.27 & 32.08 & 29.84 & 49.05 & 24.99 \\
& TRADES\cite{zhang2019theoretically} & 57.82 & 32.52 & 30.38 & 51.30 & 25.02 \\
& ARD\cite{goldblum2020adversarially} & 60.94 & 35.31 & 32.72 & 53.67 & 26.04 \\
& IAD\cite{zhu2021reliable} & 60.43 & 35.75 & 32.80 & 52.71 & 26.84 \\
& RSLAD\cite{zi2021revisiting} & 59.55 & 35.68 & 33.35 & 52.89 & \underline{27.77} \\
& AKD\cite{maroto2022benefits} & 57.84 & 34.32 & 31.98 & 51.06 & 26.06 \\
& AdaAD\cite{huang2023boosting} & \underline{62.19} & \underline{35.33} & \underline{32.52} & \underline{54.67} & 26.74 \\
& \textbf{DGAD} & \textbf{63.24} & \textbf{36.09} & \textbf{33.68} & \textbf{55.47} & \textbf{27.66} \\
&  & +1.05 & +0.76 & +1.16 & +0.80 & -0.11 \\
\midrule
\multirow{9}{*}{MN-V2} & PGD-AT\cite{madry2017towards} & 51.55 & 29.34 & 27.26 & 45.73 & 22.07 \\
& TRADES\cite{zhang2019theoretically} & 53.05 & 29.07 & 27.44 & 47.62 & 21.82 \\
& ARD\cite{goldblum2020adversarially} & 57.18 & 33.13 & 30.91 & 51.50 & 24.20 \\
& IAD\cite{zhu2021reliable} & 56.33 & 32.88 & 30.18 & 49.00 & 24.07 \\
& RSLAD\cite{zi2021revisiting} & 56.04 & 32.76 & 30.29 & 50.14 & 24.56 \\
& AKD\cite{maroto2022benefits} & 56.75 & 33.11 & 30.50 & 49.53 & 24.65 \\
& AdaAD\cite{huang2023boosting} & \underline{61.44} & \underline{34.75} & \underline{31.97} & \underline{54.21} & \underline{25.91} \\
& \textbf{DGAD} & \textbf{62.25}  & \textbf{34.90} & \textbf{32.64} & \textbf{54.54}  & \textbf{26.56} \\
& & +0.81 & +0.15 & +0.67 & +0.33 & +0.65 \\
\bottomrule
\end{tabular}
\end{table}

\begin{table}[t!]
\caption{\textbf{Evaluating on Tiny ImageNet.} The teacher model was trained with TRADES ($\lambda$=6) for 110 epochs. }
\centering
\scriptsize\addtolength{\tabcolsep}{11pt}
\begin{tabular}{c|l|c|c|c|c|c}
\toprule
Model & Method & Clean & FGSM & PGD & CW2 & AA \\
\midrule
\multirow{4}{*}{RN-18} & ARD\cite{goldblum2020adversarially} & 41.66	& 24.47	& 23.30 & 37.76 & 17.23\\
& RSLAD\cite{zi2021revisiting} & 40.83	& 23.45	& 22.58 & 37.12 & 17.05\\
& AdaAD\cite{huang2023boosting} & 47.54	& 24.22 & 22.82 & 42.79 & \textbf{17.41} \\
& \textbf{DGAD} & \textbf{47.92} & \textbf{24.42} & \textbf{23.05} & \textbf{42.81} & 17.18 \\
\bottomrule
\end{tabular}
\label{tab:tinyimagenet}
\end{table}

\textbf{CIFAR10/CIFAR100.} We compared the performance of our DGAD with other existing methods on CIFAR10 and CIFAR100 datasets, focusing particularly on the best checkpoint results against PGD attacks, as established in  \cite{huang2023boosting}.

For CIFAR10 evaluations in \cref{tab:cifar10}, our DGAD framework marks a distinct advancement in model performance. With the WideResNet-34-10 as the teacher model paired with ResNet18 as the student, DGAD has achieved a remarkable 0.83\% uplift in clean data accuracy, surpassing the already impressive teacher model's score with an overall accuracy of 87.20\%. Moreover, DGAD's fortification against attacks is evident with gains of 1.35\% against FGSM and 1.16\% against PGD attacks, showcasing a strengthened defense mechanism.

Switching to WideResNet-34-20 and MobileNetV2, DGAD achieves a notable 1.18\% increase in clean accuracy and consistent robustness gains—3.49\% against FGSM, 3.73\% on PGD, 1.74\% on CW2, and 3.44\% on AA attacks—validating its efficacy in improving both accuracy and defense in a cohesive manner. For WideResNet-34-10 with MobileNetV2, the results highlight its adaptability and strength in countering varied adversarial strategies while preserving or improving the accuracy of clean data.

For CIFAR100 in \cref{tab:cifar100}, DGAD has shown significant improvements over the AdaAD method. Specifically, for ResNet18 with WRN-34-10, there is an increase in clean accuracy by 1.05\% and a boost in robustness against the PGD attack by 1.16\%. For MobileNetV2, For the MobileNetV2 model, we observe a clean accuracy improvement of 0.81\%, alongside a 0.67\% uptick in PGD attack.

\textbf{Tiny ImageNet.} To evaluate performance on a more complex dataset, we tested DGAD on Tiny ImageNet using a PreActResNet18 teacher and a ResNet18 student model. As shown in \cref{tab:tinyimagenet}, DGAD outperforms all other methods, including ARD, RSLAD, and AdaAD, achieving the highest accuracy on both clean and adversarial examples. 

% DGAD outperforms the AdaAD baseline on the complex Tiny ImageNet dataset using a PreActResNet18 teacher and a ResNet18 student model, showcasing robustness across diverse adversarial challenges. 

\begin{table}[t!]
\caption{\textbf{Evaluating Transfer-based Attacks Using Various Surrogate Models on CIFAR10 with a ResNet18 Target Model.}} %Best results in \textbf{bold}; next-best \underline{underlined}.}
\scriptsize\addtolength{\tabcolsep}{9.5pt}
\centering
\begin{tabular}{l|c|c|c|c|c|c}
\toprule
Surrogate Model & \multicolumn{3}{c|}{ResNet34} & \multicolumn{3}{c}{VGG16} \\
\cline{1-7}
Method & FGSM & PGD & JSMA & FGSM & PGD & JSMA \\
\midrule
PGD-AT\cite{madry2017towards} & 63.05 & 60.58 & 84.90 & 64.06 & 62.78 & 85.77 \\
TRADES\cite{zhang2019theoretically} & 65.57 & 63.93 & 84.71 & 66.88 & 66.00 & 85.36 \\
ARD\cite{goldblum2020adversarially} & 65.26 & 63.20 & 86.06 & 66.64 & 65.43 & 87.03 \\
IAD\cite{zhu2021reliable} & 65.48	& 63.23	& 84.71 & 66.62 & 66.06 & 86.04 \\
RSLAD\cite{zi2021revisiting} & 65.06 & 62.77 & 85.43 & 65.91 & 64.83 & 86.26 \\
AKD\cite{maroto2022benefits} & 64.34 & 62.23 & 85.22 & 65.24 & 64.30 & 86.24 \\
AdaAD\cite{huang2023boosting} & \underline{66.81} & \underline{64.57} & \underline{88.00} & \underline{68.74} & \underline{67.89} & \underline{88.39} \\
\textbf{DGAD} & \textbf{67.77} & \textbf{65.20} & \textbf{90.56} & \textbf{70.92} & \textbf{70.29} & \textbf{90.34} \\
\bottomrule
\end{tabular}
% \vspace{-8pt}
\label{tab:surrogate_performance}
\end{table}

\begin{table}[t!]
\caption{\textbf{Evaluating Self-Adversarial Distillation on CIFAR10 using a TRADES ($\lambda=6$) trained ResNet18 Teacher Model.}}
\vspace{-5pt}
\centering
\scriptsize\addtolength{\tabcolsep}{14pt}
\begin{tabular}{l|c|c|c|c|c}
\toprule
Method & Clean & FGSM & PGD & CW2 & AA \\
\midrule
PGD-AT\cite{madry2017towards} & 82.95 & 57.16 & 52.87 & 77.56 & 47.69 \\
ARD\cite{goldblum2020adversarially} & 80.66 & 55.68 & 50.90 & 74.87 & 46.61 \\
IAD\cite{zhu2021reliable} & 81.32 & 57.54 & 52.91 & 75.69 & 48.20 \\
RSLAD\cite{zi2021revisiting} & 81.92 & 57.94 & 53.29 & 76.26 & 49.06 \\
AKD\cite{maroto2022benefits} & \textbf{83.74} & \underline{58.87} & \underline{54.17} & \textbf{77.97} & 48.84 \\
AdaAD\cite{huang2023boosting} & 83.13 & 57.54 & 53.30 & 77.62 & \underline{49.61} \\
\textbf{DGAD} & \underline{83.26} & \textbf{58.91} & \textbf{54.37} & \underline{77.90} & \textbf{50.54} \\
\bottomrule
\end{tabular}
\label{tab:rn18_performance}
\end{table}

\textbf{Transfer-based Attacks on CIFAR10.} 
We tested DGAD against transfer-based attacks using surrogate models such as ResNet34 and VGG16 on CIFAR10. Our evaluation simulates real-world scenarios, where attackers lack specific details of the target models. In \cref{tab:surrogate_performance}, DGAD outperforms all other methods, including the state-of-the-art AdaAD, across all metrics, demonstrating superior transferability and robustness against diverse surrogate model-based attacks.

\textbf{Adversarial Self-Distillation on CIFAR10.} As shown in \cref{tab:rn18_performance}, DGAD outperforms both AdaAD and AKD \cite{maroto2022benefits} in adversarial resilience, achieving superior performance against FGSM, PGD, and AA attacks. While AKD is specifically designed for self-distillation within the same architecture, DGAD demonstrates superior performance in various setups, although it may occationally lag behind AKD in scenarios tailored to AKD's design. 

\section{Conclusion}
\label{sec:conclusion}
In this study, we presented the Dynamic Guidance Adversarial Distillation (DGAD) framework, a novel strategy designed to enhance both the adversarial robustness and clean data accuracy of student models through a tailored approach in adversarial distillation. DGAD leverages Misclassification-Aware Partitioning (MAP), Error-corrective Label Swapping (ELS), and Predictive Consistency Regularization (PCR) to meticulously correct the inaccuracies in the teacher model's predictions and fine-tune the student's learning process.

Our findings affirm the effectiveness of DGAD, demonstrating substantial improvements in the model's defense against adversarial threats and its accuracy on clean data. This advancement in adversarial and knowledge distillation sets new standards for developing resilient and accurate machine learning models, paving the way for future research in enhancing model robustness without compromising performance.

% \begin{table}[tb]
%   \caption{Font sizes of headings. 
%     Table captions should always be positioned \emph{above} the tables.
%   }
%   \label{tab:headings}
%   \centering
%   \begin{tabular}{@{}lll@{}}
%     \toprule
%     Heading level & Example & Font size and style\\
%     \midrule
%     Title (centered)  & {\Large\bf Lecture Notes \dots} & 14 point, bold\\
%     1st-level heading & {\large\bf 1 Introduction} & 12 point, bold\\
%     2nd-level heading & {\bf 2.1 Printing Area} & 10 point, bold\\
%     3rd-level heading & {\bf Headings.} Text follows \dots & 10 point, bold\\
%     4th-level heading & {\it Remark.} Text follows \dots & 10 point, italic\\
%   \bottomrule
%   \end{tabular}
% \end{table}

\section*{Acknowledgements}
This work was supported by the National Research Foundation of Korea (NRF) grant funded by the Korea government (MSIT) (No. RS-2023-00222385) and partly by the Institute of Information \& communications Technology Planning \& Evaluation (IITP) grant funded by the Korea government (MSIT) (RS-2021-II212068, Artificial Intelligence Innovation Hub).

% ---- Bibliography ----
%
% BibTeX users should specify bibliography style 'splncs04'.
% References will then be sorted and formatted in the correct style.
%
\bibliographystyle{splncs04}
\bibliography{main}
\end{document}